\pdfoutput=1

\documentclass[11pt]{article}

\usepackage[final]{acl}

\usepackage{times}
\usepackage{latexsym}

\usepackage[T1]{fontenc}

\usepackage[utf8]{inputenc}

\usepackage{microtype}

\usepackage{inconsolata}

\usepackage{graphicx}

\usepackage{algorithm}
\usepackage{algorithmic}
\usepackage{amsmath}
\usepackage{amssymb} 
\usepackage{url}
\usepackage{xcolor}
\usepackage{subcaption}
\usepackage{array}
\usepackage{booktabs} 
\usepackage{multirow} 
\usepackage{makecell}
\usepackage{enumitem}
\usepackage{soul}

\definecolor{myPurple}{RGB}{128, 0, 128}

\title{MixLLM: Dynamic Routing in Mixed Large Language Models}

\author{
\textbf{Xinyuan Wang\textsuperscript{1}}\thanks{Work done during an internship at NEC Labs America.},
\textbf{Yanchi Liu\textsuperscript{2}}\thanks{Corresponding author.},
\textbf{Wei Cheng\textsuperscript{2}},
\textbf{Xujiang Zhao\textsuperscript{2}},
\\
\textbf{Zhengzhang Chen\textsuperscript{2}},
\textbf{Wenchao Yu\textsuperscript{2}},
\textbf{Yanjie Fu\textsuperscript{1}},
\textbf{Haifeng Chen\textsuperscript{2}}
\\
\\
\textsuperscript{1}Arizona State University,
\textsuperscript{2}NEC Labs America
\\
\{xwang735, yanjie.fu\}@asu.edu \\ \{yanchi, weicheng, xuzhao, zchen, wyu, haifeng\}@nec-labs.com
}

\begin{document}
\maketitle

\begin{abstract}

Large Language Models (LLMs) exhibit potential artificial generic intelligence recently, 
however, their usage is costly with high response latency. 
Given mixed LLMs with their own strengths and weaknesses, LLM routing aims to identify the most suitable model for each query in the stream to maximize response quality and minimize cost and latency.
However, the challenges involve: (1) dynamic trade-offs among quality, cost, and latency; (2) enabling continual learning in deployed systems; and 
(3) navigating a varying (e.g., new LLM addition or old LLM removal) set of LLM candidates over time.  
To bridge these gaps, we develop MixLLM, a dynamic contextual-bandit-based routing system for query-LLM assignment. 
Specifically, we first leverage query tags to enhance query embeddings for the routing task. 
Next, we design lightweight prediction models to estimate the response qualities and costs of queries over LLMs. 
We then devise a meta-decision maker to choose the query-LLM assignments to best tradeoff response quality, cost, and latency. 
Finally, the system benefits from continual training, allowing it to adapt to evolving queries and user feedback over time.
Our extensive experiments show that MixLLM achieves the best trade-offs in response quality, cost, and latency (97.25\% of GPT-4's quality at 24.18\% of the cost under the time constraint).

\end{abstract}

\section{Introduction}


Large Language Models (LLMs) have exhibited abilities to understand massive texts, generate actionable knowledge, enable contextual reasoning, and innovate diverse applications \cite{radford2018improving, radford2019language, brown2020language, raffel2020exploring, chowdhery2023palm, touvron2023llama}.
However, deploying LLMs presents unique challenges in managing computational resources, optimizing response times, and ensuring scalability. 



\begin{figure}
\centering
\includegraphics[width=0.45\textwidth]{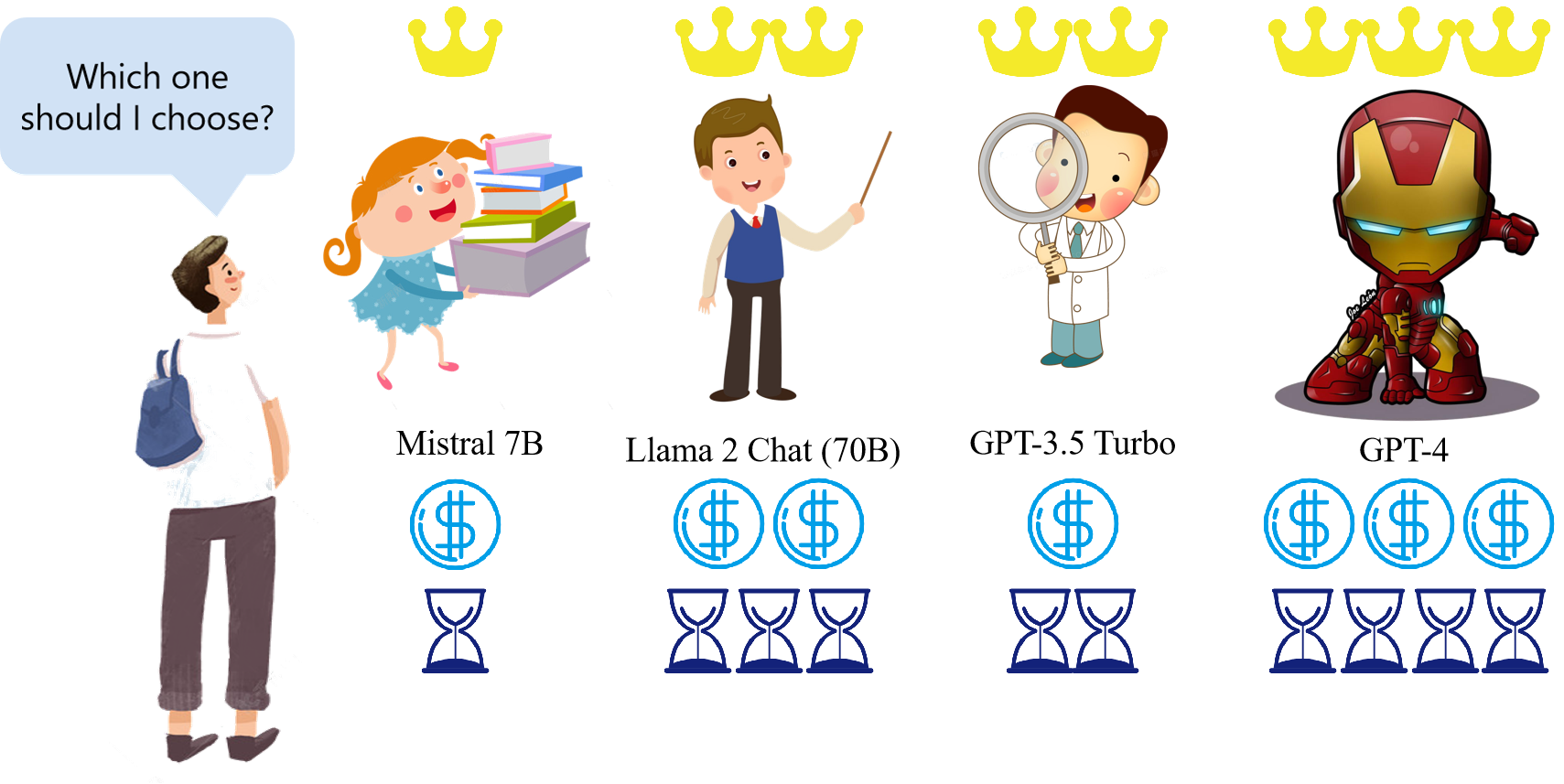}
\caption{Which is the most suitable LLM?}
\label{intro_question}
\end{figure}

As shown in \textbf{Figure \ref{intro_question}}, the diversity of available LLMs~\cite{jiang2024empowering, gong2024evolutionary, li2023sehf, li2024sade, wang2022hierarchal, wang2024llm, wang2024lcmdc}, each with different strengths and weaknesses, poses a challenge when selecting the most appropriate model for a given task. More powerful models, such as GPT-4 \cite{achiam2023gpt}, can deliver high-quality responses, but the pricy cost and computational requirements limit their accessibility. 
Thus, LLM routing, which chooses the most suitable LLMs for incoming queries in mixed LLM candidates, is needed to balance the trade-offs between response quality, cost, and latency.


Existing LLM routing methods can be categorized into non-predictive methods and predictive methods. Non-predictive methods, like cascading \cite{chen2023frugalgpt,madaan2023automix}, firstly exploit smaller LLMs and then switch to larger LLMs based on a reviewer model, but this increases both cost and latency as multiple LLMs are involved \cite{tay2022ul2}.
Predictive methods predict the performance of candidate LLMs to select the best one for each query.
For example, HybridLLM \cite{ding2024hybrid} exploits a binary classifier to predict the query difficulty for routing, 
RouterBench \cite{hu2024routerbench} predicts response quality directly, and FORC \cite{vsakota2024fly} optimizes quality and cost at the set level.

However, the challenges involved in existing work are multifaceted. 
Firstly, a key limitation of current methods is the lack of consideration 
of high latency when too many queries are routed to the same LLM. Ignoring latency can create bottlenecks, reduce system efficiency, and impact user experience. 
Secondly, continual learning in a deployed system poses a significant challenge: LLMs must adapt to evolving queries and learn from user feedback over time to maintain relevance and accuracy, necessitating robust mechanisms for incorporating feedback.
Lastly, flexibly managing the addition and removal of candidate LLMs is essential; as advancements in model architectures and techniques emerge, the routing system needs to integrate new models and retire outdated ones to ensure users benefit from the latest advancements.

To address these challenges, we propose MixLLM, a dynamic contextual-bandit-based routing system for query-LLM assignment.
First, we propose a tag-enhanced embedding model by using tags generated from the InsTag~\cite{lu2023instag} model. These tags help improve the query representations from noises.
Next, we design lightweight prediction models for each LLM to estimate response quality and cost. These LLM-specific predictions do not require system-wide retraining when new LLMs are introduced.
The meta decision-maker then selects the best LLM for each query based on the predictions. It balances trade-offs between response quality, cost, and latency to optimize query-LLM assignments.
Finally, MixLLM benefits from continual training, allowing the system adapts to evolving queries and user feedback over time, improving the performance in real-world deployment.

Our extensive experiments demonstrate that MixLLM effectively balances response quality, cost, and latency, achieving 97.25\% of GPT-4's quality at only 24.18\% of the cost. By incorporating a latency penalty, MixLLM avoids congestion and high-latency issues, ensuring efficient system performance even under heavy load. Additionally, we extend the RouterBench dataset by incorporating the latest Llama 3.1 model, showcasing the framework's scalability and adaptability. The results from online training further validate the effectiveness of the continual training approach.

Our contributions are as follows:
\begin{itemize}[nosep]
    \item We propose MixLLM that leverages enhanced query embeddings, latency penalties, and continual learning to balance response quality, cost, and latency in LLM routing.
    \item MixLLM accounts for real-world query streams by introducing a latency mechanism that factors in hardware limitations.
    \item MixLLM offers key benefits, including selecting the optimal LLM, handling the latency constraint, and adapting over time to changing environments and user feedback.
    \item We extend the RouterBench dataset by incorporating the latest Llama 3.1 model and adding prompt and response length.
\end{itemize}

\section{Related Work}
Studies on selecting the most suitable LLM include \emph{non-predictive} and \emph{predictive} routing systems.

\subsection{Non-predictive Routing System}
Non-predictive systems incorporate LLM inference during routing. A common approach is \emph{cascading}, where smaller models are used first, switching to larger ones if needed.
FrugalGPT \cite{chen2023frugalgpt} introduced three strategies to reduce cost while maintaining response quality: prompt adaptation, LLM approximation, and LLM cascade form a chain of LLMs, selecting LLMs from small to large.  
AutoMix \cite{madaan2023automix} introduced a similar cascading strategy, where a self-reviewer judges the answer and a meta-reviewer decides whether switching to a larger model is needed.
However, in non-predictive routing systems, one query may need to be answered by several LLMs which increases both cost and resource usage.

\begin{figure*}[htbp]
\centering
\includegraphics[width=0.95\linewidth]{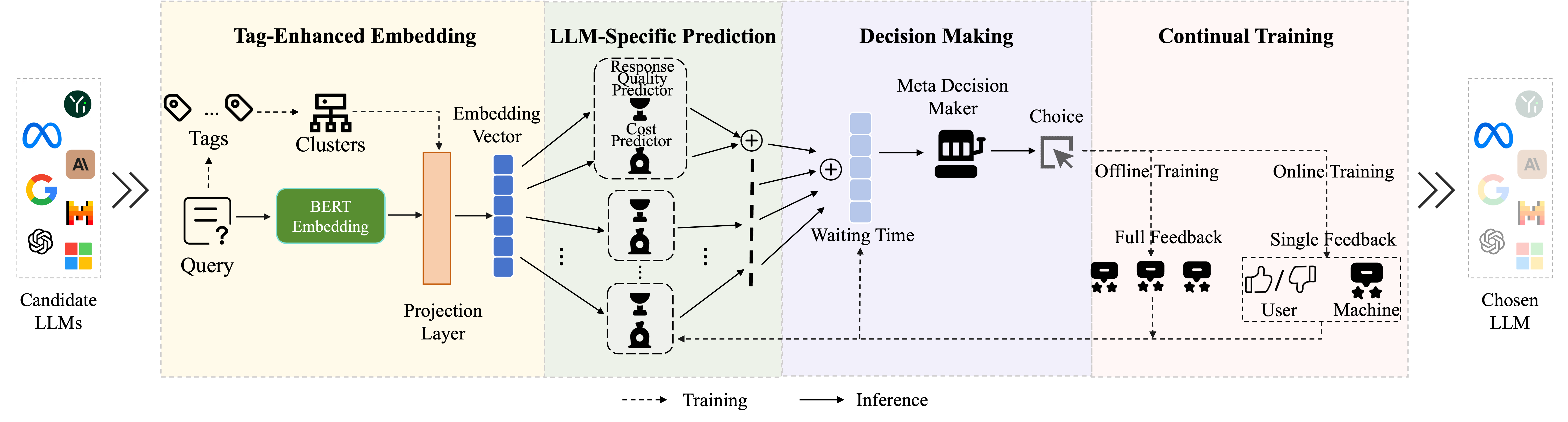}
\caption{Overview of the MixLLM Framework}
\label{method_overview}
\end{figure*}

\subsection{Predictive Routing System}
Predictive systems estimate the quality of LLM response before making routing decisions and route each query to only one LLM. These systems typically fall into categories such as classification, quality prediction, optimization, and bandit-based solutions, each offering unique strategies. Given the capability of LLMs to handle tasks across diverse domains, including medical applications~\cite{liu2024calorie,liu2025calorie}, bioinformatics~\cite{ying2024revolutionizing,liu2024pth}, material science~\cite{hu2024reinforcement}, industrial engineering~\cite{xie2025transformer,xie2024spatio}, etc, these routing systems are crucial for optimization in real-world scenarios.
\textit{Classification-based} approaches predict the best LLM for a query by treating LLMs as labels.
HybridLLM \cite{ding2024hybrid} trained a binary classifier to assign ''easy'' queries to smaller models. ME-Switch \cite{liu2024me} extended to a multi-label domain classifier, improving memory and computation efficiency. Other methods like Zooter \cite{lu-etal-2024-routing} introduced a reward model for ranking responses from different LLMs, using tag-based label enhancement for training data. RouteLLM \cite{ong2024routellm} introduced four distinct routing strategies, including similarity-weighted ranking, matrix factorization, and supervised and prompting classification. However, query labels may shift when new and powerful LLMs emerge.
\textit{Response quality prediction} methods focus on predicting the quality of each LLM's response for a given query.
Shnitzer et al.~\cite{shnitzer2023large} used 3 different ways to predict ``correctness'' (response quality) for each LLM and select the best one. RouterBench \cite{hu2024routerbench} optimized the quality-cost trade-off by a ``willingness to pay'' parameter. It also introduced a large benchmark dataset for routing tasks. However, they did not predict cost and ignored latency. 
\textit{Optimization-based} methods treat LLM routing as a set-level optimization problem. 
FORC \cite{vsakota2024fly} employed predicted response quality and cost for quality-oriented and cost-oriented linear programming strategies.
OptLLM \cite{liu2024optllm} optimized the routing problem with a multi-label classification model.
But these approaches potentially ignore low cost-effective queries.
\textit{Bandit-based} methods like MetaLLM \cite{nguyen2024metallm} adopted a single bandit approach, where the system learns to balance quality and cost trade-offs over time. However, the dependency between arms can limit scalability when adding or removing LLMs.

\section{Methodology}

\subsection{The Dynamic LLM Routing Task}
We study the problem of dynamic LLM routing with streaming queries. Given queries that arrive sequentially, our goal is to assign each query to the most appropriate LLM selected from a set of candidates to trade off response quality, cost, and latency.
Formally, let the set of streaming queries be:
$
    Q = \{q_n\}_{n=1}^{|Q|}, 
$
and the set of LLM candidates be:
$
    M = \{m_l\}_{l=1}^{|M|}.
$
The objective is to select the most suitable LLM $m_n^*$ for the query $q_n$.

\subsection{Overview of The MixLLM Framework}

\textbf{Figure~\ref{method_overview}} shows that MixLLM consists of four components: (1) tag-enhanced query embedding, (2) LLM-specific prediction, (3) meta decision maker, and (4) continual learning mechanism. This framework allows MixLLM to route queries to LLMs in a dynamic system while achieving quality-cost-latency trade-offs and continual learning with a changing LLM candidate set.

\subsection{Tag-enhanced Query Embedding via Unsupervised Fine-tuning}
A query can be seen as a token sequence, thus, its embedding can be generated using a pre-trained encoder (e.g., BERT~\cite{devlin-etal-2019-bert}):
$
    \mathbf{e}_n = \texttt{Encoder}(q_n),
$
where $\mathbf{e}_n$ represents the embedding of $n$-th query $q_n$ in a query stream.
However, such general-purpose query embeddings contain too much noises and are not tailored for LLM routing. 
To address this limitation, we propose enhancing the encoder by introducing tag knowledge, which enriches the query embeddings and improves their effectiveness for routing tasks.

\begin{figure}[htbp]
\centering
\begin{subfigure}{0.49\linewidth}
\centering
\includegraphics[width=\linewidth]{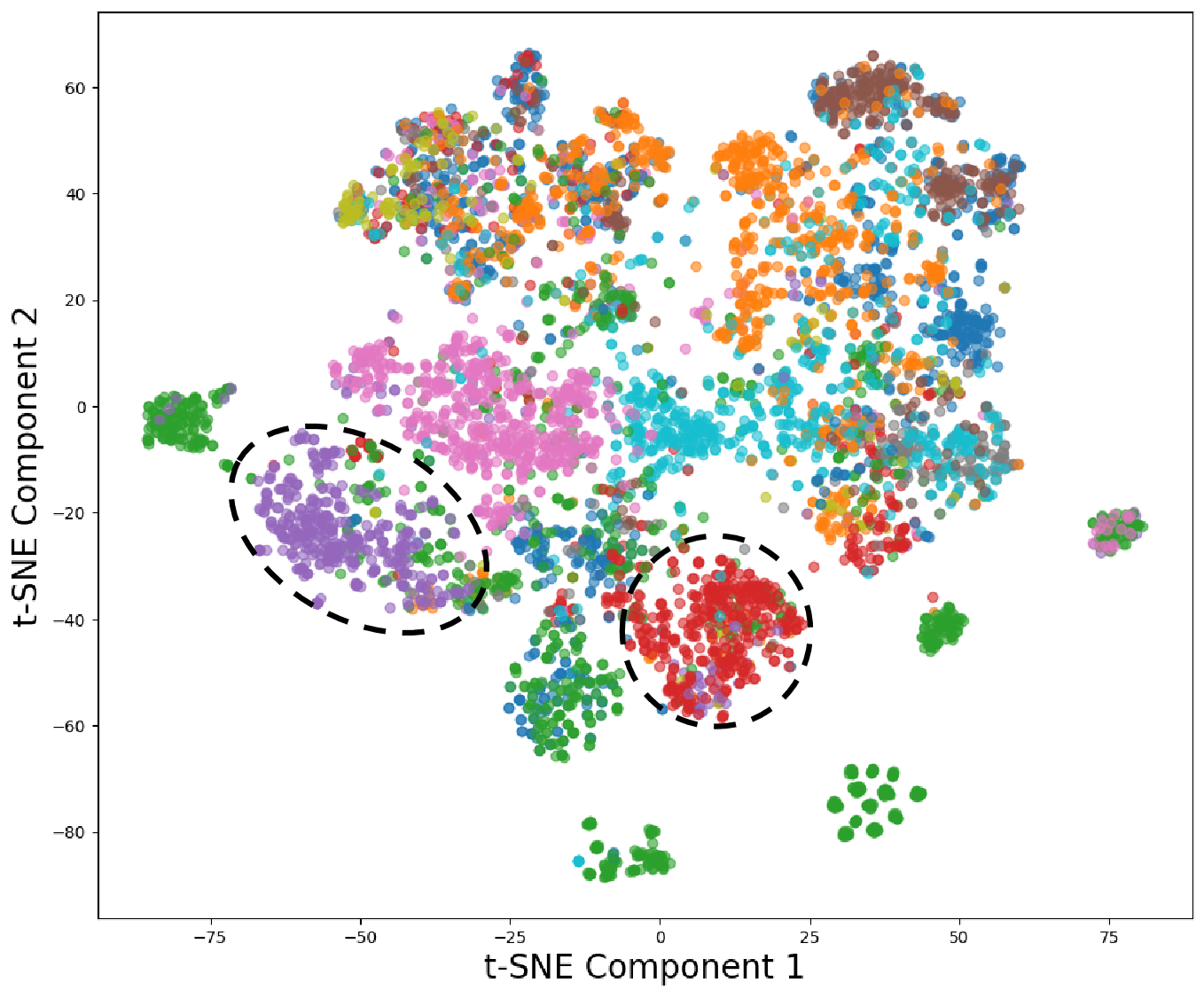}
\caption{Domain Visualization}
\label{fig:tags_a}
\end{subfigure}
\hfill
\begin{subfigure}{0.49\linewidth}
\centering
\includegraphics[width=\linewidth]{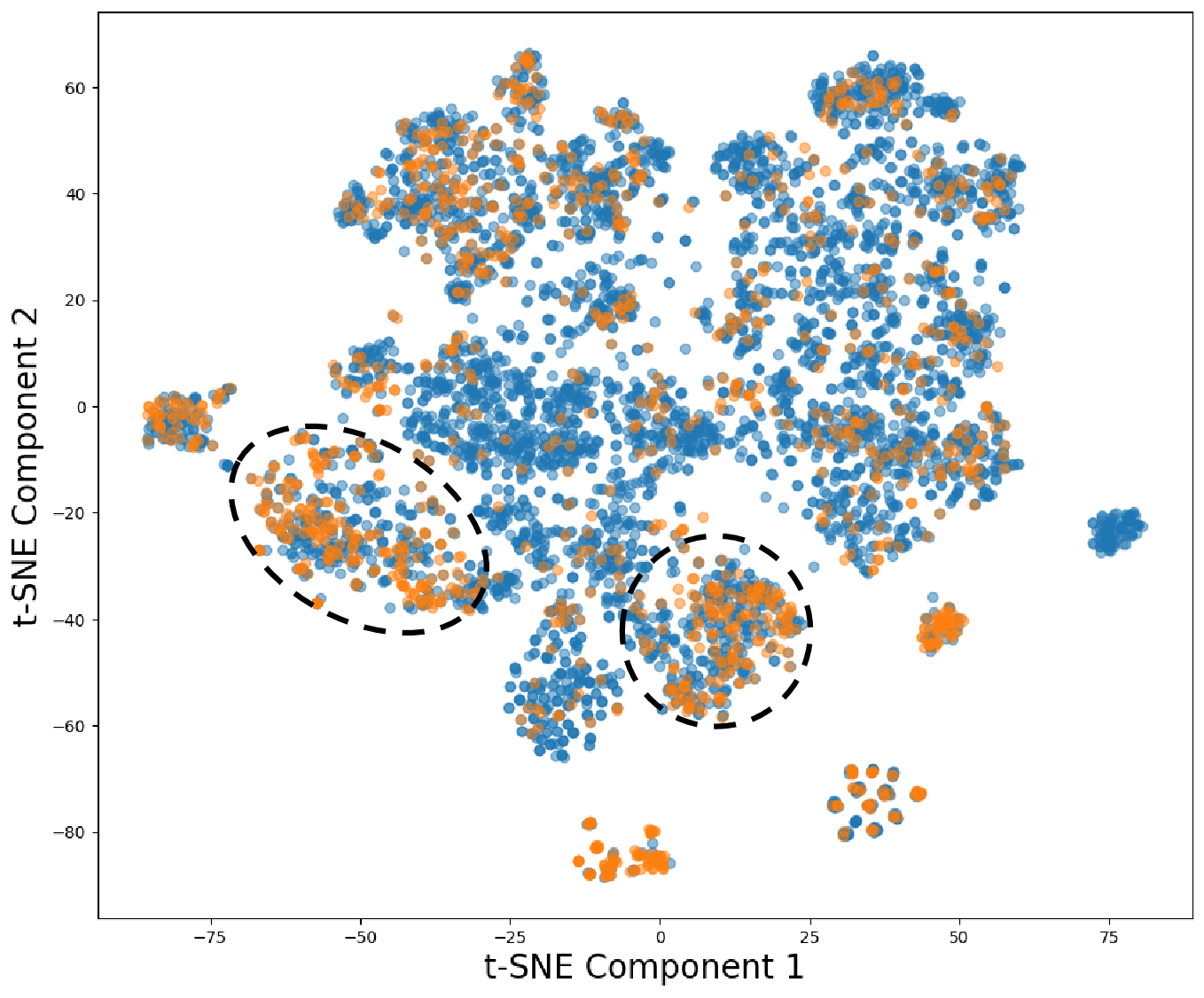}
\caption{Quality Visualization}
\label{fig:tags_b}
\end{subfigure}

\caption{Domain-Quality Correlation}
\label{fig:tags_embedding}
\end{figure}

Different LLMs can be proficient in different domains (e.g., Science, Legal)~\cite{liu-etal-2024-mathbench}.
Using GPT-4 as an example, \textbf{Figure~\ref{fig:tags_embedding}} shows a clear correlation between domain and response quality. The query distribution after t-SNE dimension reduction is shown in \textbf{Figure~\ref{fig:tags_a}}, with each color representing a specific domain. \textbf{Figure~\ref{fig:tags_b}} highlights GPT-4's response quality.
It is evident that GPT-4 has a higher error frequency (\textcolor{orange}{orange} points in \textbf{Figure~\ref{fig:tags_b}}) in the ``Legal'' (\textcolor{red}{red} points in \textbf{Figure~\ref{fig:tags_a}}) and ``Math'' (\textcolor{myPurple}{purple} points in \textbf{Figure~\ref{fig:tags_a}}) domains. 
These observations inspire us to develop the tag-enhanced embedding approach. By incorporating tags and their derived domains, we can guide embeddings to capture these distinctions, making them more suitable for LLM routing tasks.


\noindent{\bf Step 1: Automated Query Tag Generation.}
To prepare, we employ the InsTag~\cite{lu2023instag} model to generate fine-grained tags for each query and manually cluster the tags into a set of coarse-grained domains, denoted as \( D \). 
InsTag is an instruction tagging method designed to quantify the diversity and complexity of human instructions, and these tags contribute to model fine-tuning.



\noindent{\bf Step 2: Unsupervised Fine-tuning of Encoder.} 
While the InsTag model, backed by Llama-2 13B, is too large to be used during inference, we fine-tune the BERT encoder during the training stage.
We develop an unsupervised optimization objective that integrates intra-domain similarity ($\mathcal{L}_{\texttt{intra}}$) and inter-domain separation ($\mathcal{L}_{\texttt{inter}}$):
\begin{equation} \label{total_loss}
    \mathcal{L} = \mathcal{L}_{\texttt{intra}} + \mathcal{L}_{\texttt{inter}},
\end{equation}
where the intra-domain similarity loss encourages embeddings within the same domain cluster to be close to their center $\boldsymbol{\mu}_j$:
\begin{equation} \label{variation_loss}
    \mathcal{L}_{\texttt{intra}} = -\frac{1}{|Q|} \sum_{i=1}^{|Q|} \log \frac{\exp(\mathbf{e}_i \cdot \boldsymbol{\mu}_{i})}{\sum_{j=1}^{|D|} \exp(\mathbf{e}_i \cdot \boldsymbol{\mu}_j)}.
\end{equation}
The inter-domain separation loss ensures that different domain centers are distinct:
\begin{equation} \label{separation_loss}
    \mathcal{L}_{\texttt{inter}} = \frac{1}{|D|} \sum_{j=1}^{|D|} \log \sum_{k \neq j} \exp(\boldsymbol{\mu}_j \cdot \boldsymbol{\mu}_k).
\end{equation}

\subsection{LLM-Specific Quality and Cost Prediction}
Given a query embedding, we aim to predict both the response quality and financial cost for each candidate LLM on the query, so the meta decision-maker can assign the most suitable model.


\noindent\textbf{Step 1: Estimating the Response Quality of A Query-LLM Pair.} 
Since different LLMs have different response qualities, we learn an LLM-specific regression function for each LLM. This function estimates the response quality of the $n$-th query on the $l$-th LLM:
\begin{equation}
    \hat{p}_{n,l} = f_l^\texttt{rq}(\mathbf{e}_n; \boldsymbol{\theta}_l^\texttt{rq}),
\end{equation}

\noindent\textbf{Step 2: Estimating the Financial Cost of A Query-LLM Pair.} 
The total cost of the $n$-th query on the $l$-th LLM includes: 
1) the known input cost and 2) the predicted output cost, according to typical LLM pricing policies:
\begin{equation}
    \begin{aligned}
    \hat{c}_{n,l} = &\underbrace{\texttt{len}_{n,l}^{\texttt{prm}} \cdot \texttt{price}_l^{\texttt{prm}}}_{\text{input cost}} + \underbrace{\hat{\texttt{len}_{n,l}^{\texttt{res}}} \cdot \texttt{price}_l^{\texttt{res}}}_{\text{output cost}},
    \end{aligned}
\end{equation}
where $\texttt{len}_{n,l}^{\texttt{prm}}$ is the prompt length of query $q_n$, and $\texttt{price}_l^{\texttt{prm}}$ and $\texttt{price}_l^{\texttt{res}}$ are unit prices of input prompt and output response.
The response length $\hat{\texttt{len}_{n,l}^{\texttt{res}}}$ is predicted using a similar method as the response quality predictors:

\begin{equation}
    \hat{\texttt{len}_{n,l}^{\texttt{res}}} = f_l^\texttt{rl}(\mathbf{e}_n; \boldsymbol{\theta}_l^\texttt{rl}),
\end{equation}

\subsection{Meta Decision Maker}

For the $n$-th query $q_n$, the final decision score for each candidate LLM is determined by three factors: (1) $s_{n,l}^\texttt{trade}$, which trade-offs the predicted quality and cost; (2) $s_{n,l}^\texttt{unc}$, which accounts for potential prediction uncertainty; and (3) $s_{l}^\texttt{pen}$, which discourages selecting candidates with long waiting time:

\begin{equation} \label{final_scores}
    s_{n,l} = s_{n,l}^\texttt{trade} + \alpha \cdot s_{n,l}^\texttt{unc} - \beta \cdot s_{l}^\texttt{pen}.
\end{equation}
where $\alpha$ and $\beta$ control the relative importance.

The willingness to pay $\lambda$ is introduced in $s_{n,l}^\texttt{trade}$ to control the priority of quality over cost, leading to different budgets accordingly:
\begin{equation}
\label{reward_score}
    s_{n,l}^\texttt{trade} = \frac{\lambda}{\lambda + 1} \cdot \hat{p}_{n,l} - \frac{1}{\lambda + 1} \cdot \hat{c}_{n,l},
\end{equation}

To handle prediction errors, we introduce an uncertainty measurement ($s_{n,l}^\texttt{unc}$) to enhance robustness~\cite{li2010contextual}:
\begin{equation}
    s_{n,l}^\texttt{unc} = \mathbf{e}_n^T \cdot \mathbf{A}_l^{-1} \cdot \mathbf{e}_n,
\end{equation}
where \( \mathbf{A}_l^{-1} \) represents the inverse covariance matrix for the \( l \)-th LLM. This measures the amount of information gathered for each candidate and adjusts the confidence of the prediction accordingly.

Considering hardware limitations, it is crucial to avoid routing queries to candidates with excessively long waiting time. The penalty is given by:
\begin{equation}
\label{penalty_score}
    s_{l}^\texttt{pen} = e^{\gamma \cdot (w_l - \xi \cdot \tau)},
\end{equation}
where $\gamma$ is a scaling factor and $\tau$ represents the maximum tolerable waiting time. 
The waiting time $w_l$ for candidate $l$ includes: 1) the initial latency required for the LLMs to start and 2) the token output time for generating each token in the response.
The coefficient $\xi$ (smaller than 1) makes the penalty stronger. By scaling the threshold to $\xi \cdot \tau$, the system applies the penalty earlier, discouraging the selection of candidates before their waiting time reaches the full limit of $\tau$.

Finally, the candidate with the highest score is selected as the most suitable one:
\begin{equation}
    m_n^* = \arg\max_{l} (s_{n,l})
\end{equation}


\subsection{Continual Learning}
To ensure effectiveness in real-world applications, we designed both offline and online training modes. The offline mode enables the model to achieve robust performance before deployment, while the online mode allows the model to continuously improve in response to changing environments and user feedback.

\textbf{Offline Training}: 
Prior to deployment, we perform offline training using refined feedback from all candidate LLMs.  
The refined feedback includes real response quality and length, which involves updating the parameters of the predictive models:

The parameters \( \boldsymbol{\theta}_l^\texttt{rq} \) for the response quality predictors are updated using gradient descent:
\begin{equation}
    \boldsymbol{\theta}_l^\texttt{rq} := \boldsymbol{\theta}_l^\texttt{rq} - \eta_1 \cdot \nabla_{\boldsymbol{\theta}_l^\texttt{rq}} \mathcal{L}(p_{n,l}, \hat{p}_{n,l}),
    \label{rqupdate}
\end{equation}

Similarly, the response length predictor parameters \( \boldsymbol{\theta}_l^\texttt{RL} \) are updated as:
\begin{equation}
    \boldsymbol{\theta}_l^\texttt{rl} := \boldsymbol{\theta}_l^\texttt{rl} - \eta_2 \cdot \nabla_{\boldsymbol{\theta}_l^\texttt{rl}} \mathcal{L}(\texttt{len}_{n,l}^{\texttt{res}}, \hat{\texttt{len}_{n,l}^{\texttt{res}}}),
    \label{rlupdate}
\end{equation}

The uncertainty matrices \( \mathbf{A}_l \) are updated incrementally by query embeddings:
\begin{equation}
    \mathbf{A}_l := \mathbf{A}_l + \mathbf{e}_n^T \cdot \mathbf{e}_n.
    \label{uncupdate}
\end{equation}

This update accumulates information over time, decreasing the inverse \( \mathbf{A}_l^{-1} \), which leads to low uncertainty, indicating increased confidence in predictions.
Then the waiting time is adjusted based on the LLM assignment.

\textbf{Online Training}: 
After deployment, the system incrementally updates predictive models and uncertainty matrices using refined single feedback from the selected LLMs.

However, user feedback based on human satisfaction with the LLM service, often binary (``good'' or ``not good''), is challenging for training. To address this, we introduce a Dynamic Feedback Score (\( s_{n,l}^{\texttt{df}} \)) based on the contextual bandit method to capture the binary user feedback and dynamically adjust the scoring mechanism.

The final score for each LLM is updated as:
\begin{equation} \label{updated_final_scores}
    s'_{n,l} = s_{n,l} + \kappa_{n,l} \cdot s_{n,l}^{\texttt{df}},
\end{equation}
where $s_{n,l}^{\texttt{df}}$ represents the appropriateness of the $l$-th LLM to answer the given query predicted by a shared 3-layer MLP network:
\begin{equation}
    \left[ s_{n,1}^{\texttt{df}}, s_{n,2}^{\texttt{df}}, \dots, s_{n,|M|}^{\texttt{df}} \right] = f^{\texttt{df}}(\mathbf{e}_n; \boldsymbol{\theta}^{\texttt{df}}).
\end{equation}

And \( \kappa_{n,l} \) is the confidence factor based on the variance, to ensure the reliability of $s_{n,l}^{\texttt{df}}$ and prevent over-reliance on unstable predictions:
\begin{equation}
    \kappa_{n,l} = \frac{1}{\operatorname{Var}_n[s_{n,l}^{\texttt{df}}] + \epsilon},
\end{equation}
where \( \epsilon \) is a small constant to avoid division by zero. Low variance increases \( \kappa_{n,l} \), which will enhance the importance, while high variance decreases it, which reflects instability.
Then the candidate with the highest score is selected:
\begin{equation}
    m_n^* = \arg\max_{l} (s'_{n,l})
\end{equation}

Since we cannot directly supervise the network outputs with binary feedback $r_n$, we apply the \textbf{Policy Gradient} method~\cite{ban2021multi} to update \( \boldsymbol{\theta}^{\texttt{df}} \). The probability of selecting candidate \( l \) is:
\begin{equation}
    \pi(l \mid \mathbf{e}_n; \boldsymbol{\theta}^{\texttt{df}}) = \frac{\exp\left( s_{n,l}^{\texttt{df}} \right)}{\sum_{k=1}^{|M|} \exp\left( s_{n,k}^{\texttt{df}} \right)}.
\end{equation}

The goal is to maximize the expected reward:
\begin{equation}
    J(\boldsymbol{\theta}^{\texttt{df}}) = \mathbb{E}_{l \sim \pi(\cdot \mid \mathbf{e}_n; \boldsymbol{\theta}^{\texttt{df}})} [ r_n ],
\end{equation}
with gradient on selected candidate $m_n^*$:
\begin{multline}
    \nabla_{\boldsymbol{\theta}^{\texttt{df}}} \log \pi(m_n^* \mid \mathbf{e}_n; \boldsymbol{\theta}^{\texttt{df}}) = \\
    \nabla_{\boldsymbol{\theta}^{\texttt{df}}} \left( s_{n,m_n^*}^{\texttt{df}} - \log \sum_{k=1}^L \exp\left( s_{n,k}^{\texttt{df}} \right) \right)
\end{multline}

The parameters are updated as:
\begin{equation}
    \boldsymbol{\theta}^{\texttt{df}} := \boldsymbol{\theta}^{\texttt{df}} - \eta_3 \cdot \nabla_{\boldsymbol{\theta}^{\texttt{df}}} \log \pi(m_n^* \mid \mathbf{e}_n; \boldsymbol{\theta}^{\texttt{df}}) \cdot r_n.
\end{equation}

\section{Experiments}

\subsection{Experimental Settings}
\subsubsection{Dataset}
We conduct our experiments utilizing the RouterBench dataset \cite{hu2024routerbench}, which consists of 36,497 queries from 8 NLP datasets, including Chinese and English. 
Each query is answered by 11 different LLMs, with records of responses, as well as corresponding quality and cost metrics.
Moreover, we extend the dataset with Llama 3.1 8B and 70B models~\footnote{\url{https://ai.meta.com/blog/meta-llama-3-1/}} and add prompt and response lengths of all the queries and responses. The dataset is split into 80\% training and 20\% testing. 

\subsubsection{Baseline Algorithms}
We compare MixLLM with both non-predictive and predictive baselines in our experiments. 
For non-predictive methods, the cascading approach tests smaller models first and switches to larger ones if needed. We extend AutoMix~\cite{madaan2023automix} by ordering multiple LLMs by size, with cheaper models prioritized when sizes are equal.
For predictive methods, 
RouteLLM~\cite{ong2024routellm} assigns queries to LLMs using a BERT-based multi-label classifier, while  Zooter~\cite{lu-etal-2024-routing} is represented by an MLP-based classifier.
RouterBench~\cite{hu2024routerbench} predicts response quality to achieve a quality-cost trade-off.
Both FORC~\cite{vsakota2024fly} and OptLLM~\cite{liu2024optllm} predict quality and then perform set-level optimization,
while MetaLLM~\cite{nguyen2024metallm} uses a bandit algorithm with a quality-cost reward.
For additional comparison, we also include random routing and individual LLMs.

Since the baseline algorithms do not include online training after deployment, we only compare them with our offline training component for a fair comparison in Section~\ref{exp_offline}, while the online training component is further evaluated in Section~\ref{exp_online}.

\subsubsection{Evaluation Metrics}

We evaluate the methods on the streaming test queries based on the quality-cost trade-off under the latency constraint.
Specifically, the response quality score for each query is scaled from 0 to 1, while the query cost is measured in dollars. 
Any query that exceeds the maximum tolerable waiting time is assigned a quality score of 0.
The total quality and total cost are calculated as the sum of quality scores and query costs for all the test queries.
We evaluate the routing performance across varying budget levels using parameter $\lambda$, ranging from $10^{-6}$ to $10^{6}$ in Equation~\eqref{reward_score}, with a larger $\lambda$ will prioritize response quality.

\subsubsection{Configurations}

In our experiments, we set up a software environment consisting of Python 3.12, PyTorch 2.0, and CUDA 12.1 running on Ubuntu 18.04 LTS.
Most experiments were conducted on a 12GB Titan-V GPU, while tasks involving Llama models, such as dataset extension and tag extraction, were performed on two 80GB H100 GPUs.

All random seeds are set to 42 for reproducibility. In Equation~\eqref{final_scores}, $\alpha$ is set to 0.01, and $\beta$ is set to 0.1. 
In Equation~\eqref{penalty_score}, $\gamma$ is set to 0.1.

As for learning rates, $\eta_1$ and $\eta_2$ are set to 1, reflecting the use of simple machine learning algorithms, while $\eta_3$ is set to 0.001 due to the complexity of the neural network.

Query streams are configured at a rate of 100 queries per 10 seconds. The maximum tolerable waiting time $\tau$ is set to 30 seconds, and the waiting time of LLMs will be updated every 10 seconds.
The prices of input and output, the average initial time, and response speeds of different LLMs are publicly available~\footnote{\url{https://artificialanalysis.ai/}}.
This website estimates the costs of open-source LLMs based on computational resources, including CPU, GPU, and memory usage, while API-based LLMs are priced directly using their API rates.

As for quality and length regressors, we use random forest (RF) for quality prediction across all LLMs, while a combination of multi-layer perceptron (MLP), RF, and K-nearest neighbors (KNN) is applied for length (cost) prediction depending on the LLM.
Those predictors are lightweight. For example, the size of an MLP model is less than 2MB, so the inference and update time is shorter.


\begin{figure}
\centering
\includegraphics[width=0.95\linewidth]{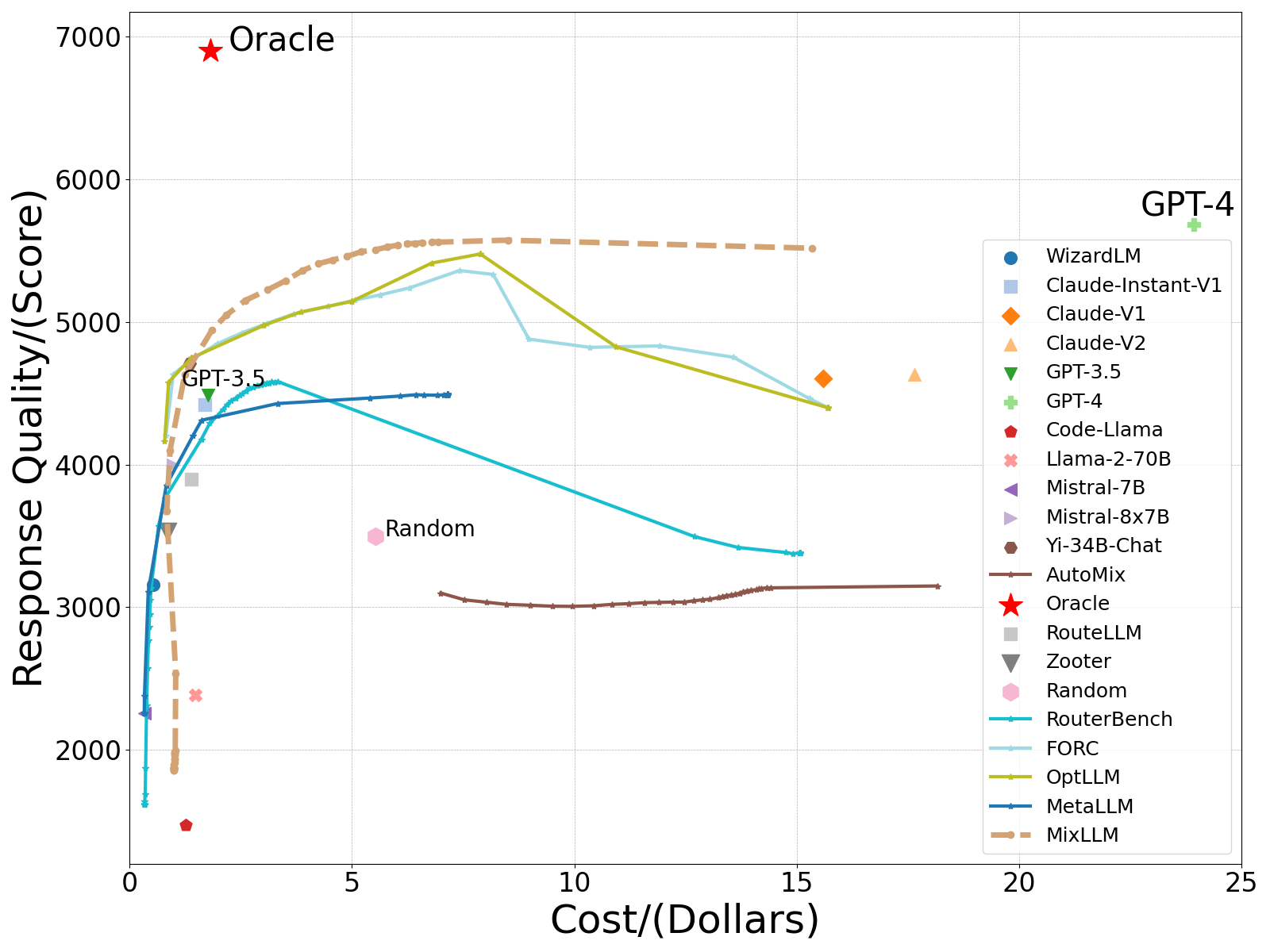}
\caption{Overall Results
}
\label{mixllm_result}
\end{figure} 

\subsection{Overall Results}
\label{exp_offline}


As shown in \textbf{Figure \ref{mixllm_result}}, MixLLM consistently outperforms the baselines, delivering strong performance. For the baseline methods, response quality can decline with larger budgets since queries may exceed the latency constraint. Notably, MixLLM achieves 97.25\% of GPT-4’s quality at only 24.18\% of the cost when $\lambda$ is 1.4. In comparison, the best baseline method, OptLLM, reaches 96.39\% of GPT-4’s quality at 32.94\% of the cost. However, beyond this point, OptLLM's response quality drops as many queries exceed the waiting time tolerance, while MixLLM remains stable. The same situation also happens on other baseline algorithms.


The \textit{Oracle} result represents the most optimal routing on this dataset, balancing response quality and cost. It serves as a benchmark for the best possible assignment. In this context, a point closer to the upper left (\textit{Oracle}) signifies higher quality at a lower cost. 
To obtain the \textit{Oracle} result, all candidate LLMs are tested for each query. For each query, the LLM that meets the quality threshold and has the lowest cost is selected. While the final results reflect only the quality and cost of the selected LLM, the process of determining the \textit{Oracle} result requires significant computational resources.

Each single LLM provides one quality-cost point. For instance, GPT-4 demonstrates superior quality, while GPT-3.5 offers a better balance of cost and quality. The ``Random'' routing serves as a baseline; points above and to the left of this anchor are superior in offering better quality at a lower cost.

AutoMix struggles because multiple LLMs handle each query, quickly exhausting the budget and reaching the latency constraint. RouteLLM and Zooter fail to adjust budgets dynamically and can only provide one quality-cost point. RouterBench performs well at lower budgets but faces latency issues as budgets increase. FORC and OptLLM share the problem of ignoring some queries due to set-level optimization, affecting user experience. MetaLLM is less effective because it can't consider multiple LLMs simultaneously, underscoring the need for a multi-armed bandit approach.






\subsection{Study on Continual Training}
\label{exp_online}

To enable continual training, we simulate the real-world query streams by splitting the training dataset into different ratios (\textbf{Table~\ref{tab:online_training_comparison}}) for offline and online training.
For example, an 80:20 split means 80\% of the data are used in offline training, while 20\% of the data are used in online training.
The offline training uses refined feedback across these splits. 
For online training, 
in addition to the refined feedback, user feedback is simulated by assuming the user is satisfied if the response quality exceeds 0.7 and the waiting time is less than 15 seconds.

\textbf{Table~\ref{tab:online_training_comparison}} presents the overall response quality for each setting, calculated as the sum of the response qualities divided by the total number of queries. 
Higher percentages indicate better performance. 
To ensure fairness, results within the same split ratio (column) maintain similar costs, which means the improvements reflect the impact of online training feedback.
Results show both types of feedback improve model performance. 
Although the improvement may seem modest, it’s important to note that online feedback is only available for the selected one, which limits the effectiveness compared to offline training. 
Despite this limitation, the results suggest that online training becomes more effective as more data are available. In real-world scenarios, where online training data are abundant, MixLLM will have greater opportunities to adapt.
Refined feedback outperforms binary feedback due to its detailed nature. Nevertheless, even the simpler binary feedback contributes to improved performance.

\begin{table}[htbp]
    \centering
    \caption{The Power of Continual Training
    }
    \resizebox{0.90\linewidth}{!}{
    \begin{tabular}{lccc}
        \toprule
        \multirow{2}{*}{\textbf{Setting}} & \multicolumn{3}{c}{\textbf{Offline : Online}} \\ 
        \cmidrule{2-4}
         & \textbf{80:20} & \textbf{50:50} & \textbf{30:70} \\ 
        \midrule
        \textbf{Without Online Training} & 75.54\% & 71.98\% & 69.74\% \\ 
        \midrule
        \textbf{With Refined Feedback} & 76.45\% & 72.99\% & 71.29\% \\ 
        \textbf{Improvement} & \textbf{1.21\%} & \textbf{1.39\%} & \textbf{2.22\%} \\ 
        \midrule
        \textbf{With Binary Feedback} & 75.93\% & 72.37\% & 70.65\% \\ 
        \textbf{Improvement} & \textbf{0.52\%} & \textbf{0.53\%} & \textbf{1.31\%} \\ 
        \bottomrule
    \end{tabular}
    }
    \label{tab:online_training_comparison}
\end{table}

In our experiments, we implemented one online test at the end of online training to demonstrate the continuing improvement of learning from and aligning with online feedback. Without loss of generality, we believe our one-time finding (online feedback can improve performance and alignment) can be generalized to recurrent tests. It is feasible to adapt our system to conduct recurrent tests at the end of each cycle in a real-world scenario.

\subsection{Study on Tag-Enhanced Embedding}

To obtain tags for the tag-enhanced encoder training, we employ InsTag~\cite{lu2023instag}, a Llama-based tagging LLM to generate one or more tags for each training query. InsTag is capable of producing over 6,000 tags, e.g., ``data structure'', ``legal ethics'', which are manually categorized into 20 domains, e.g., ``Computer Science'', ``Legal''. 


\begin{table}[htbp]
\centering
\caption{Effect of Tag-Enhanced Embedding}
\resizebox{\linewidth}{!}{
\begin{tabular}{lcccc}
\toprule
\textbf{\makecell{Cost \\ Level}} & \textbf{\makecell{Cost \\ Range}} & \textbf{\makecell{General \\ Embedding}} & \textbf{\makecell{Enhanced \\ Embedding}} & \textbf{Improvement}\\ 
\midrule
Low & $<$ 1 & 53.14\% & 56.18\% & \textbf{5.72}\% \\
Middle & 1 - 8 & 72.09\% & 73.43\% & \textbf{1.85}\% \\ 
High & $>$ 8 & 75.76\% & 76.36\% & \textbf{0.79}\% \\ 
\bottomrule
\end{tabular}
}
\label{tab:embedding_compare}
\end{table}

The results in \textbf{Table~\ref{tab:embedding_compare}} demonstrate the effectiveness of tag-enhanced embedding.
The values represent response qualities across different cost levels, where each cost level corresponds to a specific cost range. 
To ensure fairness, the costs within each level (row) are kept similar.
As the cost level increases, which corresponds to a higher budget and a greater emphasis on response quality, the improvement from tag-enhanced embedding diminishes. 
Nevertheless, at each cost level, tag-enhanced embedding consistently enhances routing performance, highlighting its importance.

\subsection{Study on Latency Constraint}

Theoretically, the trade-off between response quality and query cost often operates within the bounds of limited hardware resources in the real world. Effectively managing the workload on devices becomes essential. Different components, such as the CPU, GPU, memory, and bandwidth, all have their performance metrics, but these factors converge on one critical metric: query waiting time. Therefore, we employ the latency as the primary constraint.


We conducted a simulation to account for the latency constraint.
The total time required to answer a query has two parts: 1) the initial time to begin generating and 2) the response time, which depends on the answer length.
We use the average initial time for each LLM and estimate the response time by multiplying the output length by the corresponding LLM’s generation speed.
For closed-source LLMs, the simulation is based on API statistics. For open-source LLMs, we simulated under ideal hardware conditions, assuming sufficient memory and stable network connections to ensure optimal performance.
The average initial time and response speed of different LLMs are publicly available~\footnote{\url{https://artificialanalysis.ai/}}.

\begin{figure}[htbp]
\centering
\includegraphics[width=0.95\linewidth]{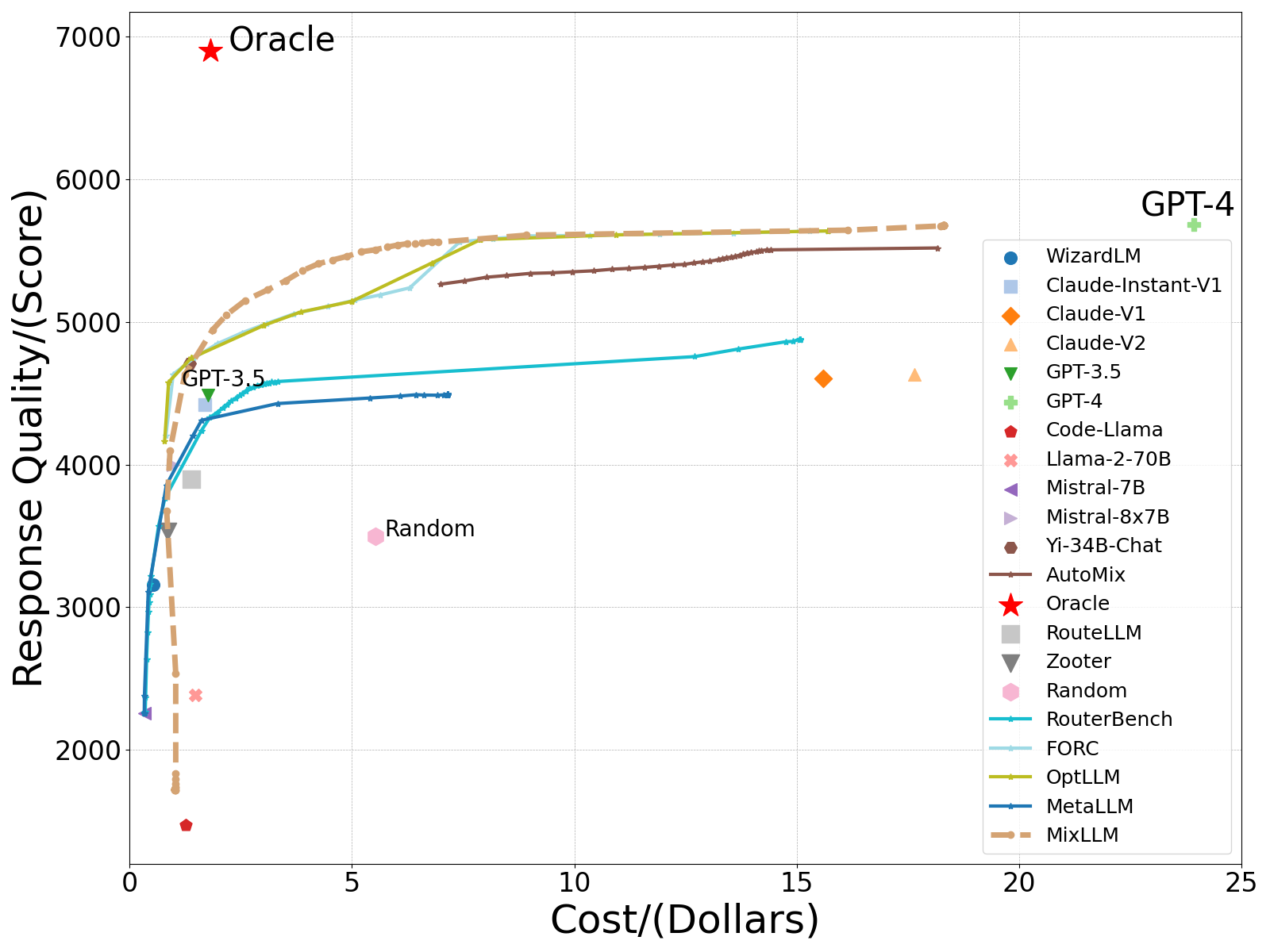}
\caption{Results without Latency Constraint}
\label{mixllm_no_latency}
\end{figure}

Even without the latency constraint, MixLLM still outperforms the baselines, as shown in \textbf{Figure~\ref{mixllm_no_latency}}. 
When compared to results with latency constraint (\textbf{Figure~\ref{mixllm_result}}), MixLLM maintains stable performance due to the time penalty component. 
However, the baselines show more variation.

In \textbf{Figure~\ref{mixllm_result}}, AutoMix's performance drops the most, primarily due to its cascading nature. Each query starts with the first LLM, resulting in significantly increased waiting time. 
Other predictive baselines also experience performance declines at higher cost levels, as they tend to route queries to more powerful LLMs with longer waiting times. This results in many queries exceeding the maximum tolerable waiting time and going unanswered.

\subsection{Study on Adaptive Training}

\begin{figure}[htbp]
\centering
\includegraphics[width=0.95\linewidth]{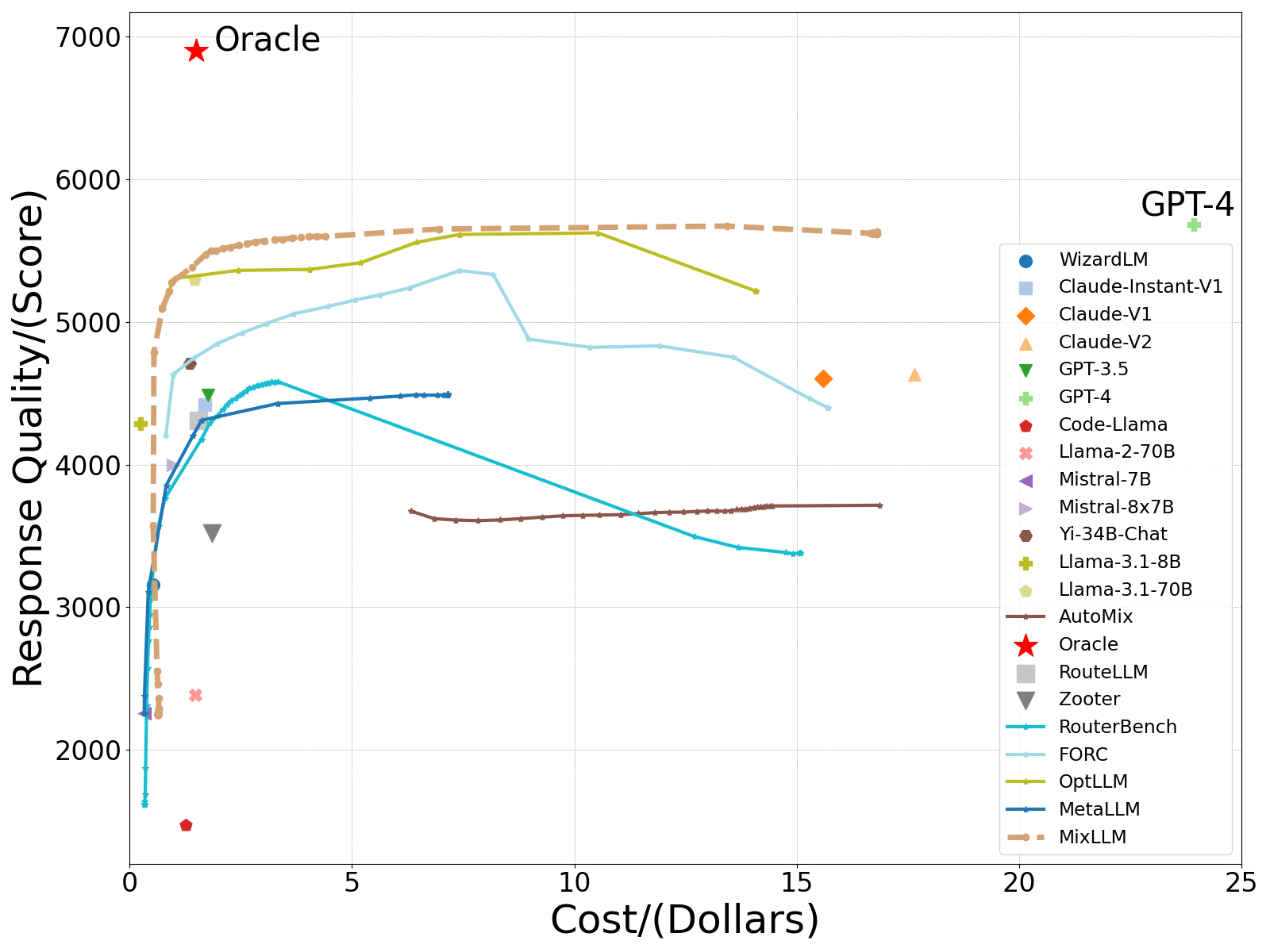}
\caption{Results with Updated Candidates}
\label{mixllm_extend}
\end{figure}

Each LLM in MixLLM operates independently, ensuring scalability. Adding or removing candidate LLMs does not require complete re-training, which only affects the corresponding LLM. To demonstrate this advantage, we extended the RouterBench dataset using new Llama 3.1 models. Specifically, we utilized the Llama 3.1 8B and 70B models to answer each query in the dataset. Then we record responses and measure their quality, cost, and length. 
As shown in \textbf{Figure~\ref{mixllm_extend}},
with the introduction of the powerful Llama 3.1 models, MixLLM achieves 98.55\% of GPT-4's response quality while reducing the cost to just 16.79\% when $\lambda$ is 1.8. Furthermore, MixLLM continues to outperform other baselines.



\subsection{Study on Out-of-Domain Generalization}
\label{sec:ood_study}
Real-world queries often originate from new or unseen domains, presenting challenges for LLM routing systems. To evaluate the domain adaptation and generalization capabilities of MixLLM, we conducted an out-of-domain (OOD) experiment. In this setup, we simulate an OOD scenario using the domains defined by tags. 
We maintain an 80:20 splitting ratio, where the testing set (20\% of the data) contains non-overlapping domains not present in the training set (80\% of the data). This design ensures that some testing samples belong to entirely unseen domains during training.

\begin{table}[htbp]
\centering
\caption{Result on OOD Scenario}
\resizebox{\linewidth}{!}{
\begin{tabular}{lcc}
\toprule
\textbf{Splitting Policy} & \textbf{Offline Only} & \textbf{Offline $+$ Online} \\ 
\midrule
Normal 80:20 Splitting & 75.54\% & 76.45\% \\
OOD 80:20 Splitting & 71.43\% & 73.89\% \\ 
Decrease & 5.44\% & \textbf{3.35}\% \\ 
\bottomrule
\end{tabular}
}
\label{tab:ood_compare}
\end{table}

The results in \textbf{Table~\ref{tab:ood_compare}} reveal that when using only offline training, MixLLM's performance decreases by 5.44\% at the same price cost level. However, when integrating both offline and online training, the performance drop is mitigated to 3.35\%. This demonstrates that the integrated offline-online training strategy effectively enhances domain generalization and adaptation. 
Furthermore, we identify MixLLM's OOD problem as a novel routing task, calling on the research community to explore and incorporate advanced domain adaptation techniques into frameworks like ours to better address this pressing challenge.

\subsection{Study on Different Choice Policy}
\label{policy_study}
\begin{figure}[htbp]
\centering
\includegraphics[width=0.95\linewidth]{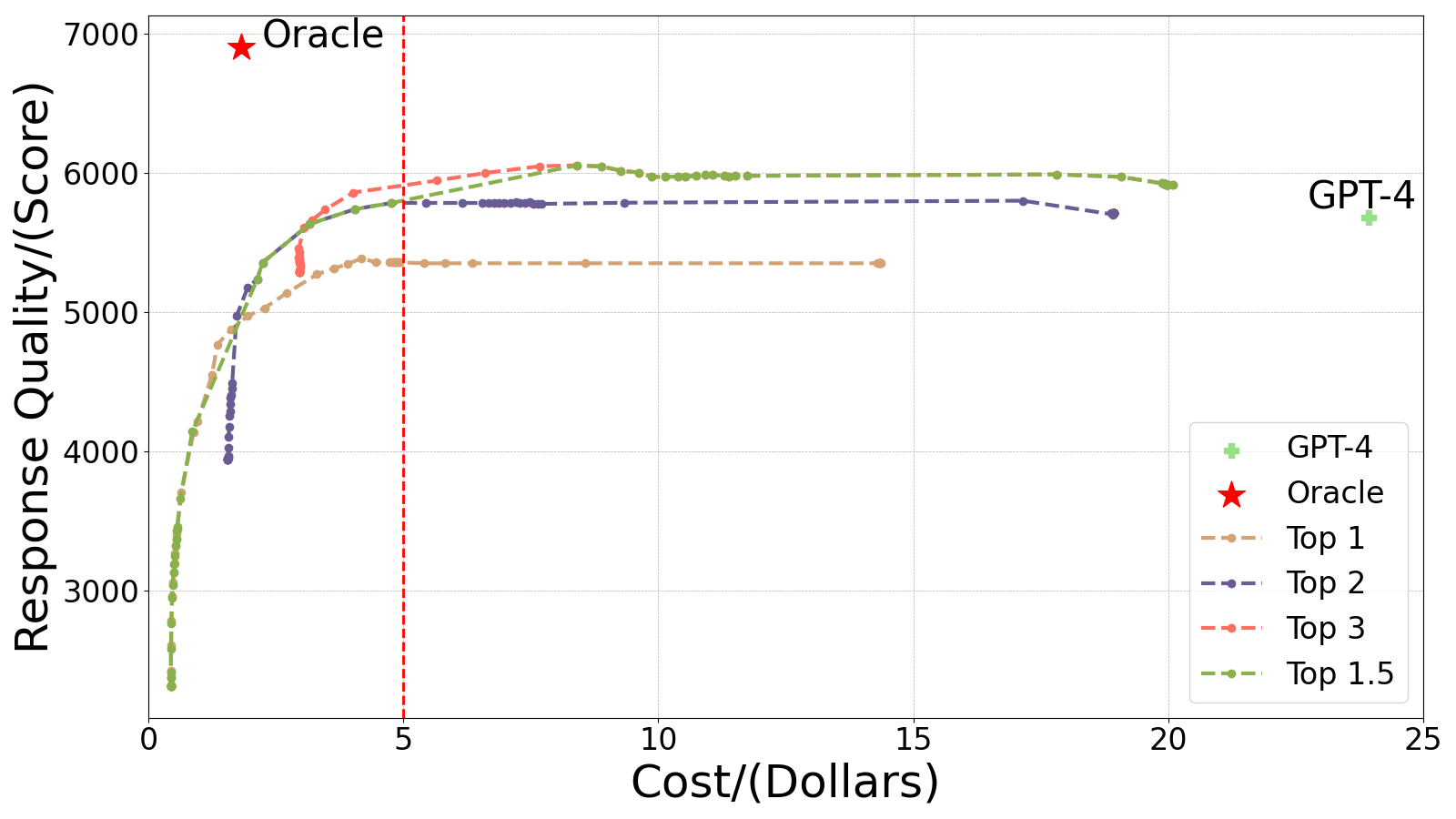}
\caption{Results on Choice Policy Study}
\label{policy_result}
\end{figure}

During our experiments, a new question arises: \ul{\textit{Can selecting more LLMs improve performance?}} To explore this, we applied various selection policies, with the results presented in \textbf{Figure~\ref{policy_result}}. 

``Top 1'', ``Top 2'', and ``Top 3'' refer to policies where the LLM(s) with the highest 1, 2, or 3 scores are selected. When multiple LLMs are chosen, the response quality reflects the best one, while costs are summed. The ``TOP 1.5'' policy introduces a dynamic adjustment, which selects the top 1 LLM when the budget is low and expands to include more LLMs as the budget increases.
As illustrated in \textbf{Figure~\ref{policy_result}}, increasing the number of selected LLMs shifts the curve upwards and to the right. This outcome is expected because selecting more LLMs increases both cost and the likelihood of choosing the most capable model. Notably, with the same budget (\textcolor{red}{red} line in \textbf{Figure~\ref{policy_result}}), the ``Top 3'' policy achieves the highest response quality, even surpassing the most powerful single LLM, GPT-4, at only 20\% of its cost.

However, in practical scenarios, users typically seek a single, definitive answer rather than multiple options. \ul{\textit{How to select the final answer?}} 
Adding a reviewer to choose the best answer is one potential solution, but it requires additional time and resources. Given the complexity, we did not incorporate a multi-choice selection into MixLLM. It presents interesting engineering challenges, and we welcome further exploration and collaboration for those interested in addressing this problem.

\section{Conclusion}


We proposed MixLLM, a dynamic routing system that selects the most suitable LLM for each query by balancing response quality, cost, and latency. By enhancing query embeddings with tag knowledge and incorporating latency constraints, MixLLM effectively addresses key challenges in real-world LLM deployment. The system's adaptability, achieved through continual learning and independent prediction for each LLM, ensures efficiency as queries evolve and new models are introduced. Our results demonstrate that MixLLM optimizes resource usage while maintaining strong performance across varying budget levels.



\section*{Limitations}

Although MixLLM presents strong performance in the experiments, some limitations are listed as follows.
(1) The training process assumes access to refined feedback, including response quality and cost, which may not always be available in real world. Training-free methods could help, such as scaling laws~\cite{ruan2024observational}.
(2) MixLLM may face challenges when routing queries from brand-new domains, commonly referred to as the out-of-domain (OOD) problem (see Section~\ref{sec:ood_study} for further details).
(3) MixLLM faces challenges in practical scenarios requiring the selection of a single definitive answer from multiple LLM outputs, as discussed in Section~\ref{policy_study}.
(4) While MixLLM considers hardware limitation through the latency constraint, more detailed dispatch strategies considering system information could further improve its practicality.
(5) More complex routing tasks remain unexplored, such as hierarchical routing. This could involve first routing a query to a relevant domain, and then selecting the most suitable LLM within that domain.
(6) MixLLM’s performance needs to be tested in real-world applications to ensure its robustness beyond idealized environments.

\section*{Acknowledgments}

Dr. Yanjie Fu is supported by the National Science Foundation (NSF) via the grant numbers: 2426340, 2416727, 2421864, 2421865, 2421803, and National Academy of Engineering Grainger Foundation Frontiers of Engineering Grants.

\bibliography{anthology,custom}

\begin{thebibliography}{39}
\providecommand{\natexlab}[1]{#1}

\bibitem[{Achiam et~al.(2023)Achiam, Adler, Agarwal, Ahmad, Akkaya, Aleman, Almeida, Altenschmidt, Altman, Anadkat et~al.}]{achiam2023gpt}
Josh Achiam, Steven Adler, Sandhini Agarwal, Lama Ahmad, Ilge Akkaya, Florencia~Leoni Aleman, Diogo Almeida, Janko Altenschmidt, Sam Altman, Shyamal Anadkat, et~al. 2023.
\newblock Gpt-4 technical report.
\newblock \emph{arXiv preprint arXiv:2303.08774}.

\bibitem[{Ban et~al.(2021)Ban, He, and Cook}]{ban2021multi}
Yikun Ban, Jingrui He, and Curtiss~B Cook. 2021.
\newblock Multi-facet contextual bandits: A neural network perspective.
\newblock In \emph{Proceedings of the 27th ACM SIGKDD Conference on Knowledge Discovery \& Data Mining}, pages 35--45.

\bibitem[{Brown et~al.(2020)Brown, Mann, Ryder, Subbiah, Kaplan, Dhariwal, Neelakantan, Shyam, Sastry, Askell et~al.}]{brown2020language}
Tom Brown, Benjamin Mann, Nick Ryder, Melanie Subbiah, Jared~D Kaplan, Prafulla Dhariwal, Arvind Neelakantan, Pranav Shyam, Girish Sastry, Amanda Askell, et~al. 2020.
\newblock Language models are few-shot learners.
\newblock \emph{Advances in neural information processing systems}, 33:1877--1901.

\bibitem[{Chen et~al.(2023)Chen, Zaharia, and Zou}]{chen2023frugalgpt}
Lingjiao Chen, Matei Zaharia, and James Zou. 2023.
\newblock Frugalgpt: How to use large language models while reducing cost and improving performance.
\newblock \emph{arXiv preprint arXiv:2305.05176}.

\bibitem[{Chowdhery et~al.(2023)Chowdhery, Narang, Devlin, Bosma, Mishra, Roberts, Barham, Chung, Sutton, Gehrmann et~al.}]{chowdhery2023palm}
Aakanksha Chowdhery, Sharan Narang, Jacob Devlin, Maarten Bosma, Gaurav Mishra, Adam Roberts, Paul Barham, Hyung~Won Chung, Charles Sutton, Sebastian Gehrmann, et~al. 2023.
\newblock Palm: Scaling language modeling with pathways.
\newblock \emph{Journal of Machine Learning Research}, 24(240):1--113.

\bibitem[{Devlin et~al.(2019)Devlin, Chang, Lee, and Toutanova}]{devlin-etal-2019-bert}
Jacob Devlin, Ming-Wei Chang, Kenton Lee, and Kristina Toutanova. 2019.
\newblock \href {https://doi.org/10.18653/v1/N19-1423} {{BERT}: Pre-training of deep bidirectional transformers for language understanding}.
\newblock In \emph{Proceedings of the 2019 Conference of the North {A}merican Chapter of the Association for Computational Linguistics: Human Language Technologies, Volume 1 (Long and Short Papers)}, pages 4171--4186, Minneapolis, Minnesota. Association for Computational Linguistics.

\bibitem[{Ding et~al.(2024)Ding, Mallick, Wang, Sim, Mukherjee, Ruhle, Lakshmanan, and Awadallah}]{ding2024hybrid}
Dujian Ding, Ankur Mallick, Chi Wang, Robert Sim, Subhabrata Mukherjee, Victor Ruhle, Laks~VS Lakshmanan, and Ahmed~Hassan Awadallah. 2024.
\newblock Hybrid llm: Cost-efficient and quality-aware query routing.
\newblock \emph{arXiv preprint arXiv:2404.14618}.

\bibitem[{Gong et~al.(2024)Gong, Reddy, Ying, and Fu}]{gong2024evolutionary}
Nanxu Gong, Chandan~K Reddy, Wangyang Ying, and Yanjie Fu. 2024.
\newblock Evolutionary large language model for automated feature transformation.
\newblock \emph{arXiv preprint arXiv:2405.16203}.

\bibitem[{Hu et~al.(2024{\natexlab{a}})Hu, Bieker, Li, Jiang, Keigwin, Ranganath, Keutzer, and Upadhyay}]{hu2024routerbench}
Qitian~Jason Hu, Jacob Bieker, Xiuyu Li, Nan Jiang, Benjamin Keigwin, Gaurav Ranganath, Kurt Keutzer, and Shriyash~Kaustubh Upadhyay. 2024{\natexlab{a}}.
\newblock Routerbench: A benchmark for multi-llm routing system.
\newblock \emph{arXiv preprint arXiv:2403.12031}.

\bibitem[{Hu et~al.(2024{\natexlab{b}})Hu, Wang, Ying, and Fu}]{hu2024reinforcement}
Xuanming Hu, Dongjie Wang, Wangyang Ying, and Yanjie Fu. 2024{\natexlab{b}}.
\newblock Reinforcement feature transformation for polymer property performance prediction.
\newblock In \emph{Proceedings of the 33rd ACM International Conference on Information and Knowledge Management}, pages 4538--4545.

\bibitem[{Jiang et~al.(2024)Jiang, Pan, Zhang, Garg, Schneider, Nevmyvaka, and Song}]{jiang2024empowering}
Yushan Jiang, Zijie Pan, Xikun Zhang, Sahil Garg, Anderson Schneider, Yuriy Nevmyvaka, and Dongjin Song. 2024.
\newblock Empowering time series analysis with large language models: A survey.
\newblock \emph{arXiv preprint arXiv:2402.03182}.

\bibitem[{Li et~al.(2023)Li, Peng, Wang, Mou, and Wang}]{li2023sehf}
Haozhou Li, Qinke Peng, Xinyuan Wang, Xu~Mou, and Yonghao Wang. 2023.
\newblock Sehf: A summary-enhanced hierarchical framework for financial report sentiment analysis.
\newblock \emph{IEEE Transactions on Computational Social Systems}.

\bibitem[{Li et~al.(2024)Li, Wang, Du, Sun, and Peng}]{li2024sade}
Haozhou Li, Xinyuan Wang, Hongkai Du, Wentong Sun, and Qinke Peng. 2024.
\newblock Sade: A speaker-aware dual encoding model based on diagbert for medical triage and pre-diagnosis.
\newblock In \emph{ICASSP 2024-2024 IEEE International Conference on Acoustics, Speech and Signal Processing (ICASSP)}, pages 12712--12716. IEEE.

\bibitem[{Li et~al.(2010)Li, Chu, Langford, and Schapire}]{li2010contextual}
Lihong Li, Wei Chu, John Langford, and Robert~E Schapire. 2010.
\newblock A contextual-bandit approach to personalized news article recommendation.
\newblock In \emph{Proceedings of the 19th international conference on World wide web}, pages 661--670.

\bibitem[{Liu et~al.(2024{\natexlab{a}})Liu, Liu, and Rosen}]{liu2024pth}
Hanghang Liu, Linyi Liu, and Clifford~J Rosen. 2024{\natexlab{a}}.
\newblock Pth and the regulation of mesenchymal cells within the bone marrow niche.
\newblock \emph{Cells}, 13(5):406.

\bibitem[{Liu et~al.(2024{\natexlab{b}})Liu, Zheng, Qiao, Duan, Fei, Zhou, Zhang, Zhang, Lin, and Chen}]{liu-etal-2024-mathbench}
Hongwei Liu, Zilong Zheng, Yuxuan Qiao, Haodong Duan, Zhiwei Fei, Fengzhe Zhou, Wenwei Zhang, Songyang Zhang, Dahua Lin, and Kai Chen. 2024{\natexlab{b}}.
\newblock \href {https://aclanthology.org/2024.findings-acl.411} {{M}ath{B}ench: Evaluating the theory and application proficiency of {LLM}s with a hierarchical mathematics benchmark}.
\newblock In \emph{Findings of the Association for Computational Linguistics ACL 2024}, pages 6884--6915, Bangkok, Thailand and virtual meeting. Association for Computational Linguistics.

\bibitem[{Liu et~al.(2024{\natexlab{c}})Liu, Gong, Zhang, He, Cai, and Zhuang}]{liu2024me}
Jing Liu, Ruihao Gong, Mingyang Zhang, Yefei He, Jianfei Cai, and Bohan Zhuang. 2024{\natexlab{c}}.
\newblock Me-switch: A memory-efficient expert switching framework for large language models.
\newblock \emph{arXiv preprint arXiv:2406.09041}.

\bibitem[{Liu et~al.(2025)Liu, Le, DeMambro, Feng, Liu, Ying, Baron, and Rosen}]{liu2025calorie}
Linyi Liu, Phuong~T Le, Victoria~E DeMambro, Tiange Feng, Hanghang Liu, Wangyang Ying, Roland Baron, and Clifford~J Rosen. 2025.
\newblock Calorie restriction induces mandible bone loss by regulating mitochondrial function.
\newblock \emph{Bone}, 190:117326.

\bibitem[{Liu et~al.(2024{\natexlab{d}})Liu, Le, Stohn, Liu, Ying, Baron, and Rosen}]{liu2024calorie}
Linyi Liu, Phuong~T Le, J~Patrizia Stohn, Hanghang Liu, Wangyang Ying, Roland Baron, and Clifford~J Rosen. 2024{\natexlab{d}}.
\newblock Calorie restriction in mice impairs cortical but not trabecular peak bone mass by suppressing bone remodeling.
\newblock \emph{Journal of Bone and Mineral Research}, 39(8):1188--1199.

\bibitem[{Liu et~al.(2024{\natexlab{e}})Liu, Zhang, Miao, Le, and Li}]{liu2024optllm}
Yueyue Liu, Hongyu Zhang, Yuantian Miao, Van-Hoang Le, and Zhiqiang Li. 2024{\natexlab{e}}.
\newblock Optllm: Optimal assignment of queries to large language models.
\newblock \emph{arXiv preprint arXiv:2405.15130}.

\bibitem[{Lu et~al.(2024)Lu, Yuan, Lin, Lin, Yuan, Zhou, and Zhou}]{lu-etal-2024-routing}
Keming Lu, Hongyi Yuan, Runji Lin, Junyang Lin, Zheng Yuan, Chang Zhou, and Jingren Zhou. 2024.
\newblock \href {https://doi.org/10.18653/v1/2024.naacl-long.109} {Routing to the expert: Efficient reward-guided ensemble of large language models}.
\newblock In \emph{Proceedings of the 2024 Conference of the North American Chapter of the Association for Computational Linguistics: Human Language Technologies (Volume 1: Long Papers)}, pages 1964--1974, Mexico City, Mexico. Association for Computational Linguistics.

\bibitem[{Lu et~al.(2023)Lu, Yuan, Yuan, Lin, Lin, Tan, Zhou, and Zhou}]{lu2023instag}
Keming Lu, Hongyi Yuan, Zheng Yuan, Runji Lin, Junyang Lin, Chuanqi Tan, Chang Zhou, and Jingren Zhou. 2023.
\newblock \# instag: Instruction tagging for analyzing supervised fine-tuning of large language models.
\newblock In \emph{The Twelfth International Conference on Learning Representations}.

\bibitem[{Madaan et~al.(2023)Madaan, Aggarwal, Anand, Potharaju, Mishra, Zhou, Gupta, Rajagopal, Kappaganthu, Yang et~al.}]{madaan2023automix}
Aman Madaan, Pranjal Aggarwal, Ankit Anand, Srividya~Pranavi Potharaju, Swaroop Mishra, Pei Zhou, Aditya Gupta, Dheeraj Rajagopal, Karthik Kappaganthu, Yiming Yang, et~al. 2023.
\newblock Automix: Automatically mixing language models.
\newblock \emph{arXiv preprint arXiv:2310.12963}.

\bibitem[{Nguyen et~al.(2024)Nguyen, Hoang, Decugis, Manchanda, Chawla, and Doan}]{nguyen2024metallm}
Quang~H Nguyen, Duy~C Hoang, Juliette Decugis, Saurav Manchanda, Nitesh~V Chawla, and Khoa~D Doan. 2024.
\newblock Metallm: A high-performant and cost-efficient dynamic framework for wrapping llms.
\newblock \emph{arXiv preprint arXiv:2407.10834}.

\bibitem[{Ong et~al.(2024)Ong, Almahairi, Wu, Chiang, Wu, Gonzalez, Kadous, and Stoica}]{ong2024routellm}
Isaac Ong, Amjad Almahairi, Vincent Wu, Wei-Lin Chiang, Tianhao Wu, Joseph~E Gonzalez, M~Waleed Kadous, and Ion Stoica. 2024.
\newblock Routellm: Learning to route llms with preference data.
\newblock \emph{arXiv preprint arXiv:2406.18665}.

\bibitem[{Radford et~al.(2018)Radford, Narasimhan, Salimans, Sutskever et~al.}]{radford2018improving}
Alec Radford, Karthik Narasimhan, Tim Salimans, Ilya Sutskever, et~al. 2018.
\newblock Improving language understanding by generative pre-training.

\bibitem[{Radford et~al.(2019)Radford, Wu, Child, Luan, Amodei, Sutskever et~al.}]{radford2019language}
Alec Radford, Jeffrey Wu, Rewon Child, David Luan, Dario Amodei, Ilya Sutskever, et~al. 2019.
\newblock Language models are unsupervised multitask learners.
\newblock \emph{OpenAI blog}, 1(8):9.

\bibitem[{Raffel et~al.(2020)Raffel, Shazeer, Roberts, Lee, Narang, Matena, Zhou, Li, and Liu}]{raffel2020exploring}
Colin Raffel, Noam Shazeer, Adam Roberts, Katherine Lee, Sharan Narang, Michael Matena, Yanqi Zhou, Wei Li, and Peter~J Liu. 2020.
\newblock Exploring the limits of transfer learning with a unified text-to-text transformer.
\newblock \emph{Journal of machine learning research}, 21(140):1--67.

\bibitem[{Ruan et~al.(2024)Ruan, Maddison, and Hashimoto}]{ruan2024observational}
Yangjun Ruan, Chris~J Maddison, and Tatsunori Hashimoto. 2024.
\newblock Observational scaling laws and the predictability of language model performance.
\newblock \emph{arXiv preprint arXiv:2405.10938}.

\bibitem[{{\v{S}}akota et~al.(2024){\v{S}}akota, Peyrard, and West}]{vsakota2024fly}
Marija {\v{S}}akota, Maxime Peyrard, and Robert West. 2024.
\newblock Fly-swat or cannon? cost-effective language model choice via meta-modeling.
\newblock In \emph{Proceedings of the 17th ACM International Conference on Web Search and Data Mining}, pages 606--615.

\bibitem[{Shnitzer et~al.(2023)Shnitzer, Ou, Silva, Soule, Sun, Solomon, Thompson, and Yurochkin}]{shnitzer2023large}
Tal Shnitzer, Anthony Ou, M{\'\i}rian Silva, Kate Soule, Yuekai Sun, Justin Solomon, Neil Thompson, and Mikhail Yurochkin. 2023.
\newblock Large language model routing with benchmark datasets.
\newblock \emph{arXiv preprint arXiv:2309.15789}.

\bibitem[{Tay et~al.(2022)Tay, Dehghani, Tran, Garcia, Wei, Wang, Chung, Shakeri, Bahri, Schuster et~al.}]{tay2022ul2}
Yi~Tay, Mostafa Dehghani, Vinh~Q Tran, Xavier Garcia, Jason Wei, Xuezhi Wang, Hyung~Won Chung, Siamak Shakeri, Dara Bahri, Tal Schuster, et~al. 2022.
\newblock Ul2: Unifying language learning paradigms.
\newblock \emph{arXiv preprint arXiv:2205.05131}.

\bibitem[{Touvron et~al.(2023)Touvron, Lavril, Izacard, Martinet, Lachaux, Lacroix, Rozi{\`e}re, Goyal, Hambro, Azhar et~al.}]{touvron2023llama}
Hugo Touvron, Thibaut Lavril, Gautier Izacard, Xavier Martinet, Marie-Anne Lachaux, Timoth{\'e}e Lacroix, Baptiste Rozi{\`e}re, Naman Goyal, Eric Hambro, Faisal Azhar, et~al. 2023.
\newblock Llama: Open and efficient foundation language models.
\newblock \emph{arXiv preprint arXiv:2302.13971}.

\bibitem[{Wang et~al.(2024{\natexlab{a}})Wang, Li, Zheng, and Peng}]{wang2024lcmdc}
Xinyuan Wang, Haozhou Li, Dingfang Zheng, and Qinke Peng. 2024{\natexlab{a}}.
\newblock Lcmdc: Large-scale chinese medical dialogue corpora for automatic triage and medical consultation.
\newblock \emph{arXiv preprint arXiv:2410.03521}.

\bibitem[{Wang et~al.(2022)Wang, Peng, Mou, Li, and Wang}]{wang2022hierarchal}
Xinyuan Wang, Qinke Peng, Xu~Mou, Haozhou Li, and Ying Wang. 2022.
\newblock A hierarchal bert structure for native speaker writing detection.
\newblock In \emph{2022 China Automation Congress (CAC)}, pages 3705--3710. IEEE.

\bibitem[{Wang et~al.(2024{\natexlab{b}})Wang, Wu, Hong, Liu, and Fu}]{wang2024llm}
Xinyuan Wang, Liang Wu, Liangjie Hong, Hao Liu, and Yanjie Fu. 2024{\natexlab{b}}.
\newblock Llm-enhanced user-item interactions: Leveraging edge information for optimized recommendations.
\newblock \emph{arXiv preprint arXiv:2402.09617}.

\bibitem[{Xie et~al.(2024)Xie, Hoskins, Rowe, and Ju}]{xie2024spatio}
Haoyang Xie, Dylan Hoskins, Kyle Rowe, and Feng Ju. 2024.
\newblock Spatio-temporal transformer for temperature profiles prediction in large format additive manufacturing.
\newblock In \emph{2024 IEEE 20th International Conference on Automation Science and Engineering (CASE)}, pages 1331--1336. IEEE.

\bibitem[{Xie et~al.(2025)Xie, Hoskins, Rowe, and Ju}]{xie2025transformer}
Haoyang Xie, Dylan Hoskins, Kyle Rowe, and Feng Ju. 2025.
\newblock Transformer-based offline printing strategy design for large format additive manufacturing.
\newblock \emph{Journal of Computing and Information Science in Engineering}, 25(2).

\bibitem[{Ying et~al.(2024)Ying, Wang, Hu, Qiu, Park, and Fu}]{ying2024revolutionizing}
Wangyang Ying, Dongjie Wang, Xuanming Hu, Ji~Qiu, Jin Park, and Yanjie Fu. 2024.
\newblock Revolutionizing biomarker discovery: Leveraging generative ai for bio-knowledge-embedded continuous space exploration.
\newblock In \emph{Proceedings of the 33rd ACM International Conference on Information and Knowledge Management}, pages 5046--5053.

\end{thebibliography}




\end{document}